\documentclass[conference]{IEEEtran}
\IEEEoverridecommandlockouts
\usepackage{times,amsmath,epsfig,amssymb,graphicx,amsfonts,pxfonts,txfonts}
\usepackage{subfigure}
\usepackage{psfrag}
\usepackage{ifpdf}
\usepackage{cite}
\usepackage{tikz}
\usetikzlibrary{shapes,arrows}
\usepackage{multirow}
\usepackage{url}
\usepackage{rotating}

\title{Robust Head Pose Estimation Using\\
Contourlet Transform}

\author{\IEEEauthorblockN{Mohammad Tofighi, Hashem Kalbkhani, Mahrokh G. Shayesteh, and Mehdi Ghasemzadeh}\\
\IEEEauthorblockA{Department of Electrical Engineering, Urmia University, Urmia, Iran\\
\IEEEauthorblockA{Emails: mo.tofighi@gmail.com, \{st\_h.kalbkhani, m.shayesteh\}@urmia.ac.ir, me\_ghasemzadeh@yahoo.com}
}}

\begin{document}

\pagestyle{plain}

\maketitle
\thispagestyle{plain}

\begin{abstract}
Head pose estimation is an important pre-processing step in many pattern recognition and computer vision systems such as face recognition. Since the performance of the face recognition systems is greatly affected by the pose of the face, how to estimate the accurate pose of the face in the face image is still a challenging problem. In this paper, we present a novel method for head pose estimation. To enhance the efficiency of the estimation we first use contourlet transform for feature extraction. Contourlet transform is a multi-resolution, multi-direction transform. Finally, in order to reduce the feature space dimension and obtain appropriate features, we use LDA (Linear Discriminant Analysis) and PCA (Principal Component Analysis) to remove inefficient features. Then, we apply k-nearest neighborhood (k-NN) and minimum distance classifiers to classify the pose of head. We use the public available FERET database to evaluate the performance of the proposed method. Simulation results indicate the efficiency of the proposed method in comparison with previous method.
\end{abstract}

\pagenumbering{arabic}

% NOTE keywords are not used for conference papers so do not populate them
\begin{keywords}
Contourlet transform, Head pose estimation, k-NN, LDA, PCA.
\end{keywords}

\section{Introduction}\label{introduction}
Head pose estimation is the process of extracting the orientation of a face or head from an image containing the face. Pose estimation is always used by human on relations to show a heightened sense of awareness and understanding by non-verbal communications when interacting with other people. There has been a significant improvement in face recognition over the last two decades. However, robust and accurate face recognition is still a classic problem because of variations in the pose of the face. Therefore, pose invariant face perception has been an active research topic for several years.

Several methods were introduced in the literature in the late 1990's and it continues up to now \cite{murphy}. The proposed methods had advantages and also many  disadvantages, and to overcome the disadvantages and enhance the advantages, many studies have been done in this field. The introduced methods can be classified into four categories as: 1) Template matching \cite{ba} methods which use the nearest neighbor classification to find the most similar view of a new head pose, 2) Appearance-based methods \cite{osadchy} that apply pattern classification or nonlinear regression tools, develop a functional mapping from the image or feature data to a head pose measurement, 3) Geometric methods \cite{sung} that determine poses by using the relative configuration of facial landmarks (such as eyes, mouth, nose tip), 4) Dimensionality reduction methods \cite{yan,balasubramanian,liu} which seek a low-dimensional continuous manifold constrained by the pose variations, and then new images can be embedded into these manifolds and used for template matching or regression.

In this paper, we develop a new method to estimate the pose of the head in the face image. In our proposed method, we use contourlet transform to extract the features from the image. Moreover, to reduce the dimension of the feature space and also, to increase the separation we utilize Principal Component Analysis (PCA) and Linear Discriminant Analysis (LDA). In the final step, we apply the extracted features to the classifiers such as k-nearest neighborhood (k-NN) and minimum distance to classify them.

The rest of the paper is organized as follows: Section \ref{Preliminaries} introduces the contourlet transform. Section \ref{Proposed} explains the proposed method. Simulation results and comparisons with other pose estimator systems are presented in section \ref{Simulation}. Finally, section \ref{Conlusion} concludes the paper.

\section{Contourlet Transform}\label{Preliminaries}
Minh $et\ al.$ \cite{do} developed Contourlet transform based on an efficient two-dimensional (2-D) multiscale and directional filter bank that can deal effectively with images having smooth contours. Contourlets not only possess the main features of wavelets (namely multiscale and time-frequency localization), but also offer a high degree of directionality and anisotropy. The main difference between the contourlets and the other multiscale directional systems is that the contourlet transform allows for different and flexible number of directions at each scale, while achieving nearly critical sampling rate.

\begin{figure}[h!]
\centering
\subfigure[]
{\label{fig:Contourlet3}\includegraphics[width=90mm]{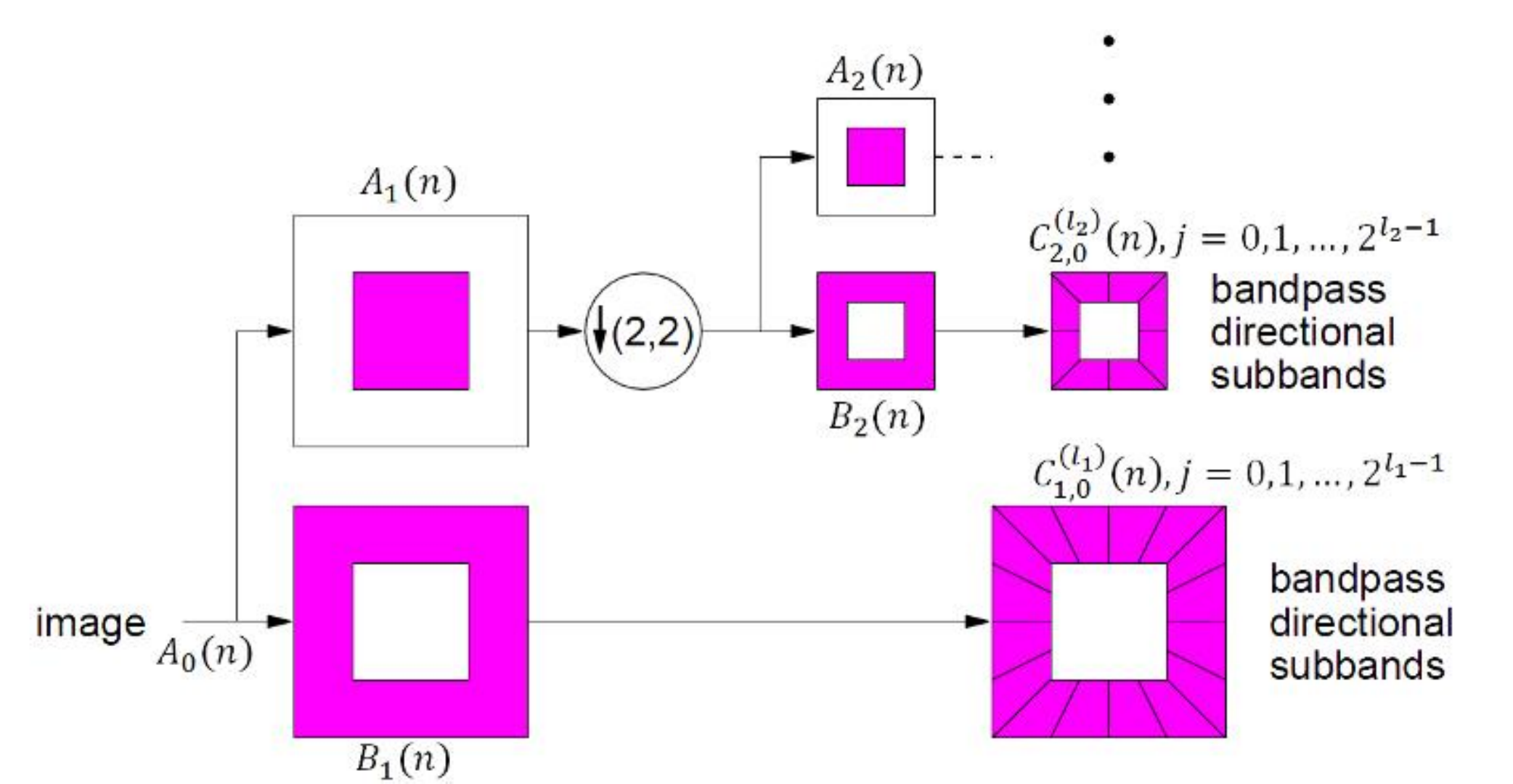}}
\subfigure[]
{\label{fig:Untitled}\includegraphics[width=50mm]{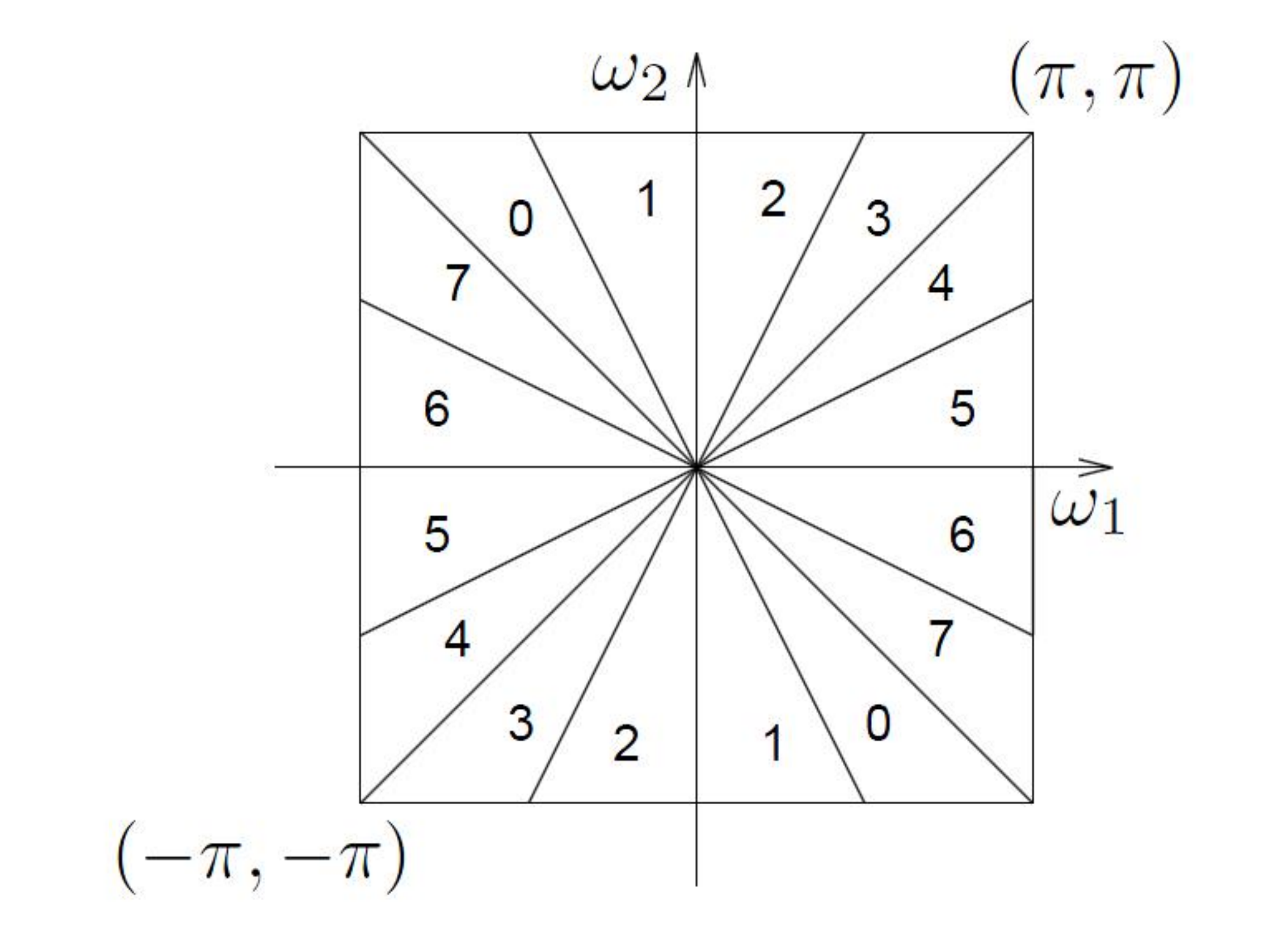}}
\caption{Spectrum partitioning, (a) decomposition scheme of the contourlet transform, (b) 8-subband partition.}
\label{fig:contourlet}
\end{figure}

In contourlet transform to achieve a sparse series for images with smooth contours and having useful computations, the iterated filter banks are utilized. Contourlet transform is composed of Laplacian Pyramid (LP) \cite{burt,vetterli} and Directional Filter Bank (DFB) \cite{bamberger}. The first stage in this transform is to capture the points discontinuities by Laplacian Pyramid, and then followed by a directional filter bank to relate point discontinuities into linear structures. At each level, the LP decomposition generates a downsampled lowpass version of the original image, and then a bandpass image is resulted from the difference between the original image and the prediction one. Subsequently, the bandpass signal is filtered in different directions. The singular points in the same direction are synthesized to a coefficient by DFB.

As shown in Fig. \ref{fig:Contourlet3}, DFB divides the highpass image into directional subbands in the frequency domain. Fig. \ref{fig:Contourlet3} shows a multiresolution and directional decomposition of contourlet filter bank which is composed of an LP and a DFB. To capture the directional information, bandpass images from the LP are fed into a DFB. Fig. \ref{fig:Untitled} shows the DFB with level $l = 3$ where there are $2^3 = 8$ real wedge-shaped frequency bands.  Subbands 0-3 correspond to the mostly horizontal directions, while subbands 4-7 correspond to the mostly vertical directions.

The DFB divides the highpass image into $2^l$ wedge-shaped frequency bands after $l$ level tree-structure segmentation. Since the DFB was designed to capture the high frequency of the input image, the low frequency content is poorly handled. In fact, the DFB alone does not provide a sparse representation for images, because with the frequency partition shown in Fig. \ref{fig:Untitled}, low frequency would leak into several directional subbands. This fact provides another reason to combine the DFB with a multiresolution decomposition, where low frequencies of the input image are removed before applying the DFB.

Let $A_0(n)$ be the input image as shown in Fig. \ref{fig:Contourlet3}. The output after the LP decomposition is a lowpass image $A_i(n)$ and $I$ bandpass images $B_i(n)$, $i=1,2, ...,I$ . That means, the $i$-th level of LP decomposes the image $A_{i-1}(n)$ into a coarser image $A_i(n)$ and a highpass image $B_i(n)$. Then each of highpass images $B_i(n)$ are further decomposed by a $l_i$-level DFB into $2^{l_i}$ highpass directional images $C$${x \atop y}, j=0,1,...,2^{l_i-1}$. The combined result is a double iterated filter bank structure, named contourlet filter bank, that is resulted in contourlet transform.

\section{Proposed Head Pose Estimation Method}\label{Proposed}
Fig. \ref{fig:flow} presents the diagram that shows the sequence we used in our proposed head pose estimation algorithm. In order to compute contourlet transform of head image, we use the contourlet transform toolbox provided in \cite{toolbox}.  In our method, at first, color face image is converted to gray-scale image. To do this, RGB image is converted to gray-scale by averaging the $r,\ g$,\ and $b$ components of the color image. This method eliminates the hue and saturation information while retaining the luminance. We use pyramidal directional filter bank (PDFB) decomposition; \textquoteleft 9-7\textquoteright\ filters for the multiscale decomposition stage and \textquoteleft PKVA\textquoteright\ filters for the directional decomposition stage \cite{po, see}. There will be two levels of pyramidal decomposition and the numbers of directional decomposition at each pyramidal level are set to 0 and 1. Using this decomposition, we obtain six sub-bands. These sub-bands represent our feature space. Fig. \ref{fig:sub} shows a typical image and the sub-bands obtained from the contourlet transform.

% Define block styles
\tikzstyle{block} = [rectangle, draw=blue, fill=black!10,
    text width=16em, text centered, rounded corners, minimum height=2em]
\tikzstyle{line} = [draw, -latex']

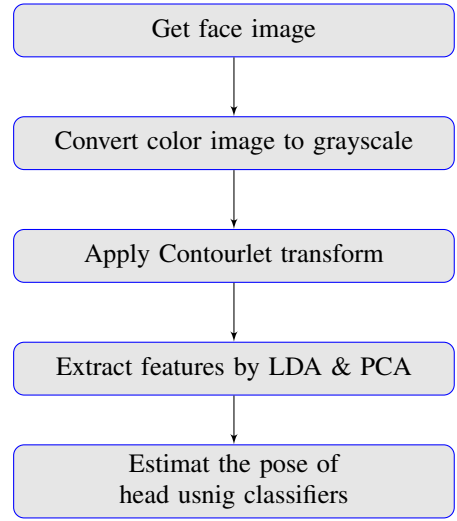
\begin{figure}[h!]
\centering
\begin{tikzpicture}[node distance = 1.5cm, auto]
    % Place nodes
    \node [block] (init) {Get face image};
    \node [block, below of=init] (identify) {Convert color image to grayscale};
    \node [block, below of=identify] (evaluate) {Apply Contourlet transform};
    \node [block, below of=evaluate] (decide) {Extract features by LDA \& PCA};
    \node [block, below of=decide] (stop) {Estimat the pose of head usnig classifiers};
    % Draw edges
    \path [line] (init) -- (identify);
    \path [line] (identify) -- (evaluate);
    \path [line] (evaluate) -- (decide);
    \path [line] (decide) -- (stop);
\end{tikzpicture}
\caption{Block diagram of the proposed head pose estimation algorithm}
\label{fig:flow}
\end{figure}

All the extracted features are not appropriate for our work; consequently, the improper features should be removed from the feature space. There are several approaches for feature space dimension reduction, such as PCA and LDA. If we use only LDA, sometimes scatter matrix in LDA becomes singular and this degrades the performance of system. Hence, we use PCA before LDA to overcome this problem. PCA is unsupervised approach which is used only for whitening. LDA is supervised algorithm, thus it is good approach for feature space dimension reduction. More details are presented in the experimental results section.

After obtaining the best feature vector, the final step is classifying the features. In this work, we choose two different classifiers: minimum distance and k-nearest neighborhood (k-NN). More details are included in the next section.

\begin{figure}[t]
\centering
\subfigure[$pl$]
{\label{fig:pl}\includegraphics[width=20mm]{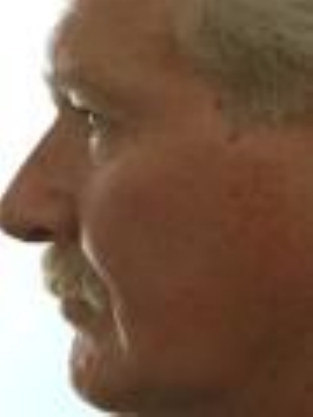}}
~\subfigure[$hl$]
{\label{fig:hl}\includegraphics[width=20mm]{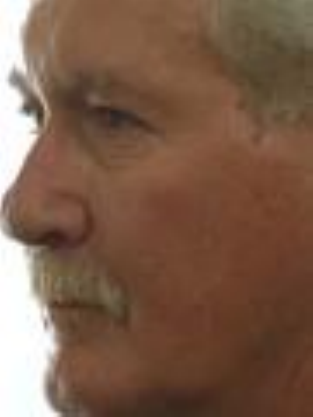}}
~\subfigure[$ql$]
{\label{fig:ql}\includegraphics[width=20mm]{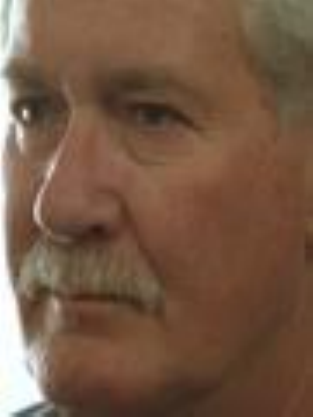}}
~\subfigure[$fa$]
{\label{fig:fa}\includegraphics[width=20mm]{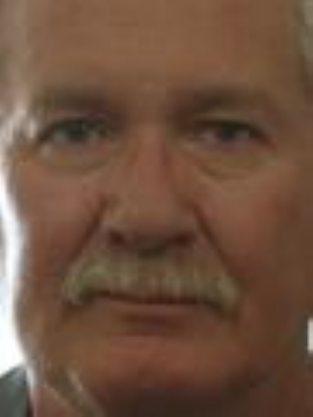}}
~\subfigure[$qr$]
{\label{fig:qr}\includegraphics[width=20mm]{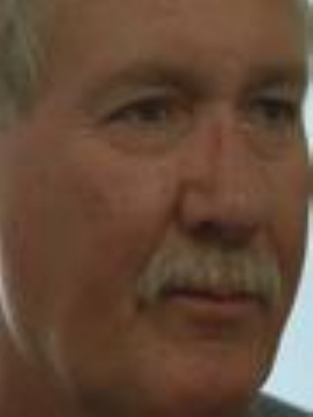}}
~\subfigure[$hr$]
{\label{fig:hr}\includegraphics[width=20mm]{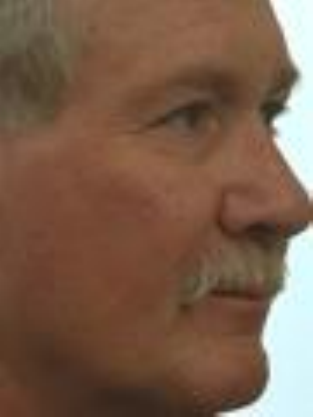}}
~\subfigure[$pr$]
{\label{fig:pr}\includegraphics[width=20mm]{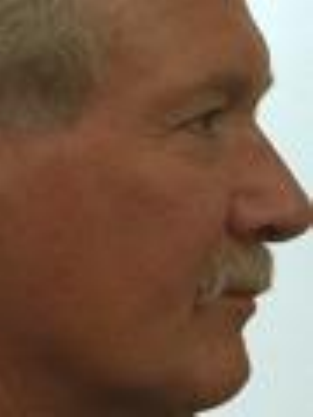}}
\caption{Different head poses with label in FERET database}
\label{fig:pose}
\end{figure}

\section{Simulation Results and Discussion}\label{Simulation}
\subsection{Database}
We need the database that contains face images with different rotations. So, we choose FERET database \cite{FERET}. In this database, lighting conditions, face size and descent of people vary too much. The images have different indices. From different indices, we select face rotations represented by $pl$,\ $hl$,\ $ql$,\ $fa$,\ $fb$,\ $qr$,\ $hr$,\ and $pr$ indices, which stand for profile left (head turned about 90 degree left), half left (head turned about 67.5 degrees left), quarter left (head turned about 22.5 degrees left), regular frontal image, alternative frontal image, taken shortly after the corresponding fa image, quarter right (head turned about 22.5 degrees right), half right (head turned about 67.5 degrees right), and profile right (head turned about 90 degree right), respectively. All images have size of $512\times768$ pixels and are in PPM (Portable Pixel Map) format. Besides, the selected images are with and without glasses. Fig. \ref{fig:pose} shows some images from FERET database.

\begin{figure}[h]
\centering
\subfigure[]
{\label{fig:a}\includegraphics[width=25mm]{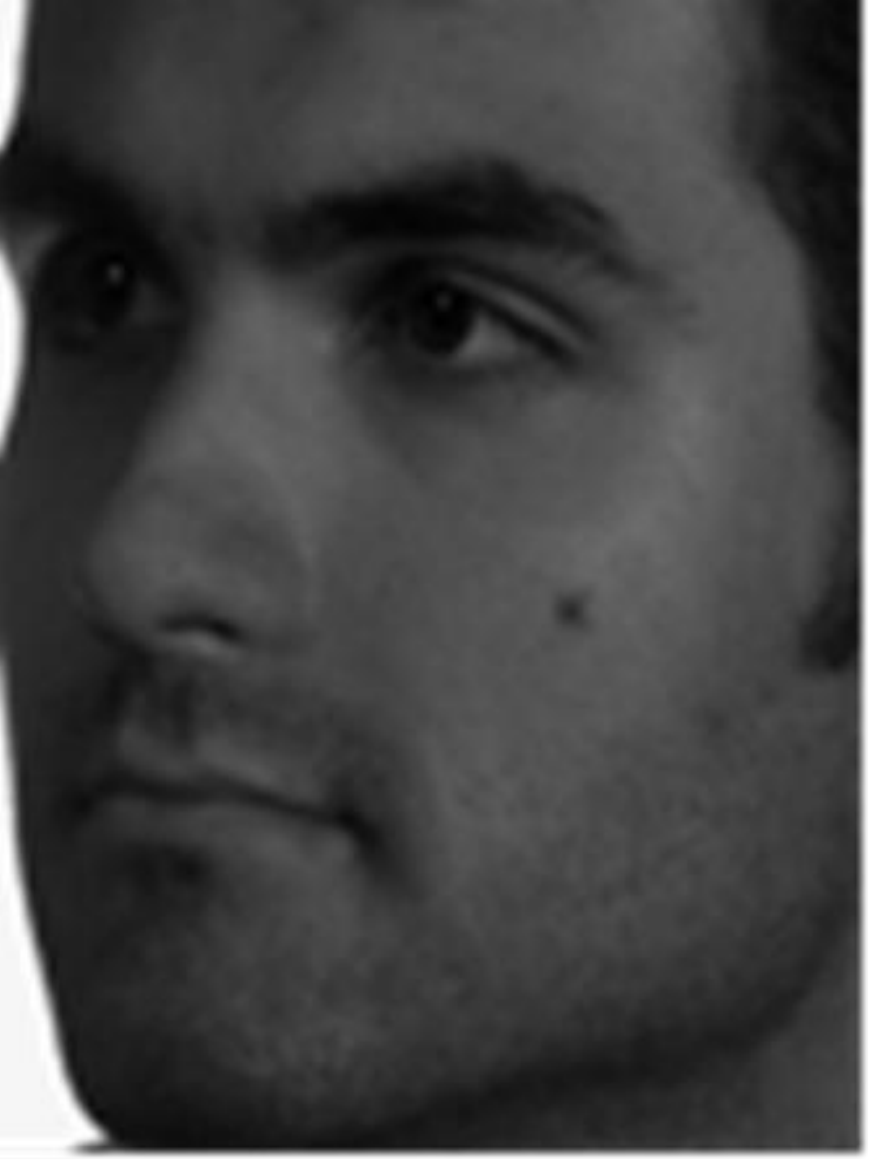}} \\
~\subfigure[]
{\label{fig:b}\includegraphics[width=8mm]{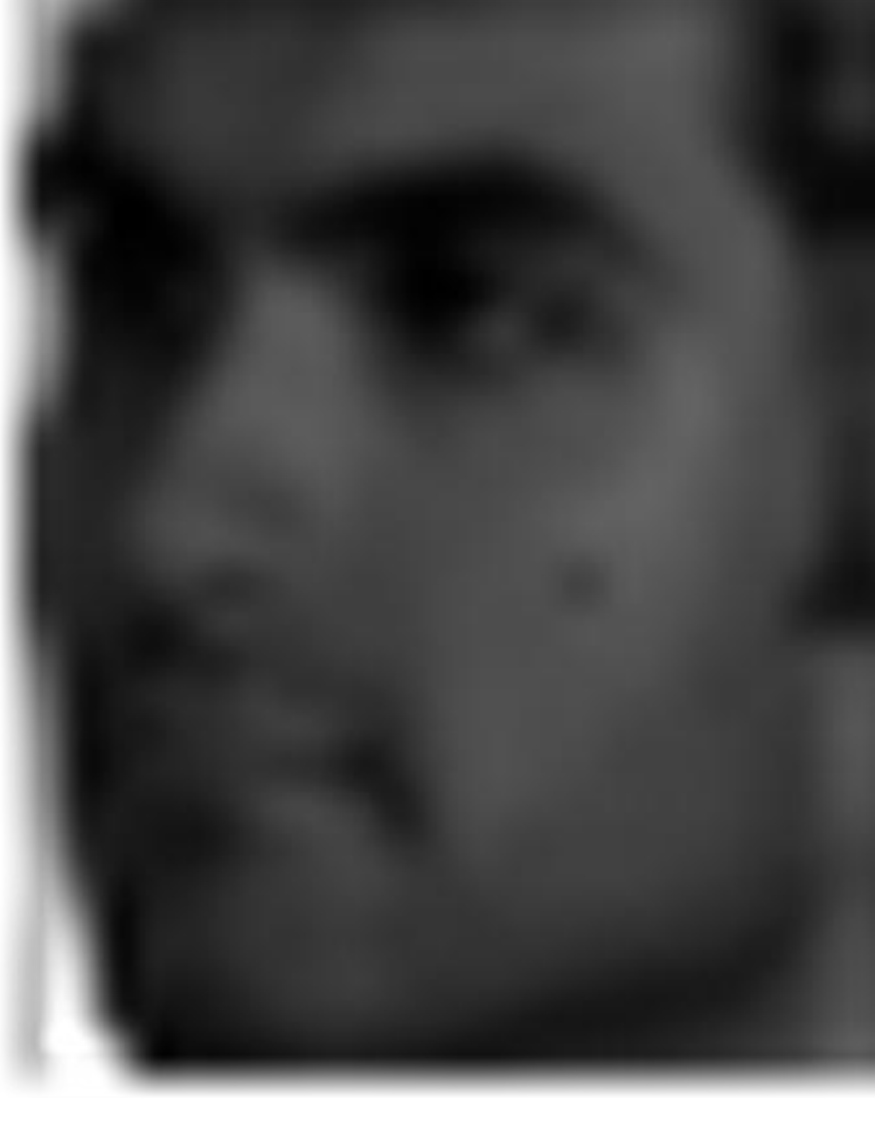}}
~\subfigure[]
{\label{fig:c}\includegraphics[width=8mm]{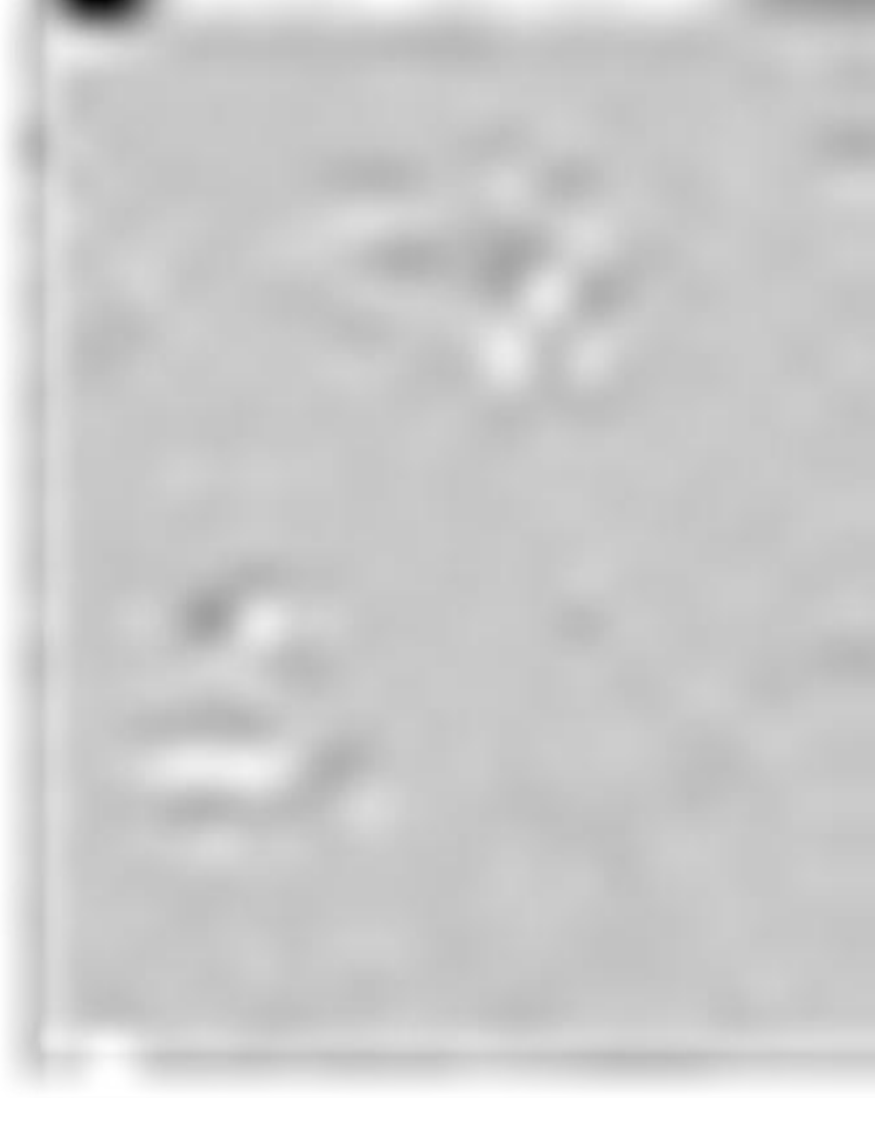}}
~\subfigure[]
{\label{fig:d}\includegraphics[width=25mm]{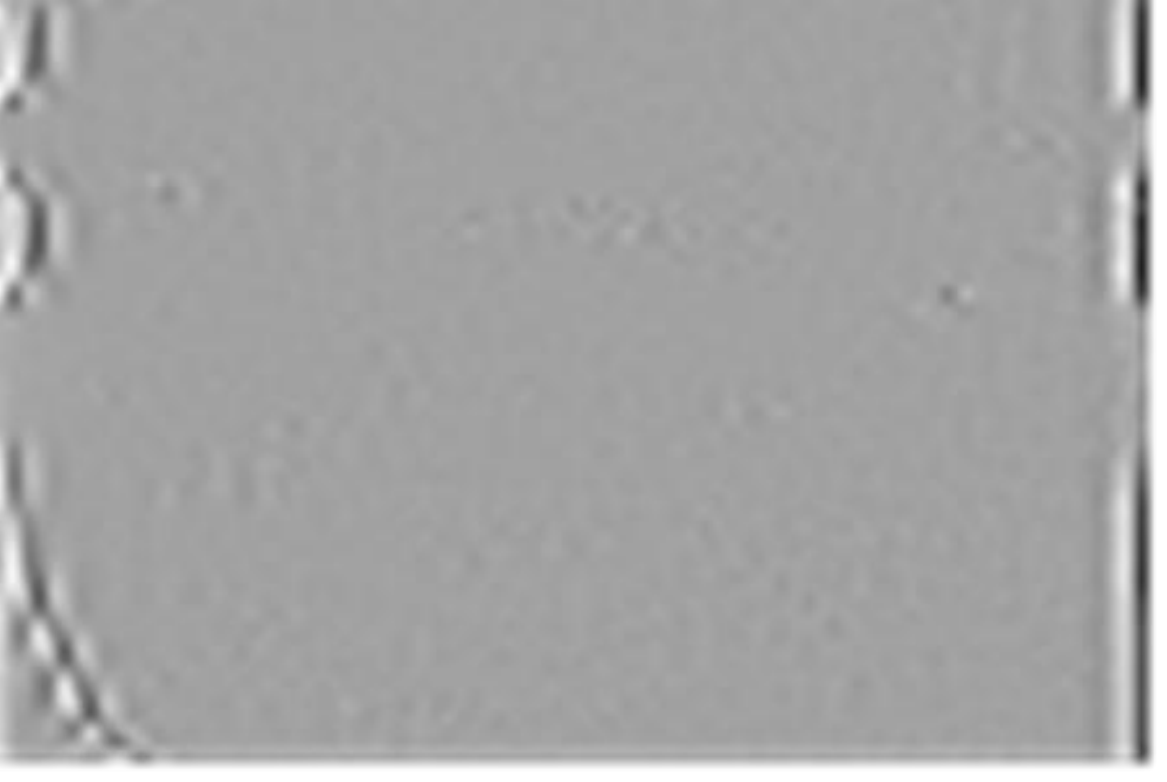}} \\
~\subfigure[]
{\label{fig:e}\includegraphics[width=8mm]{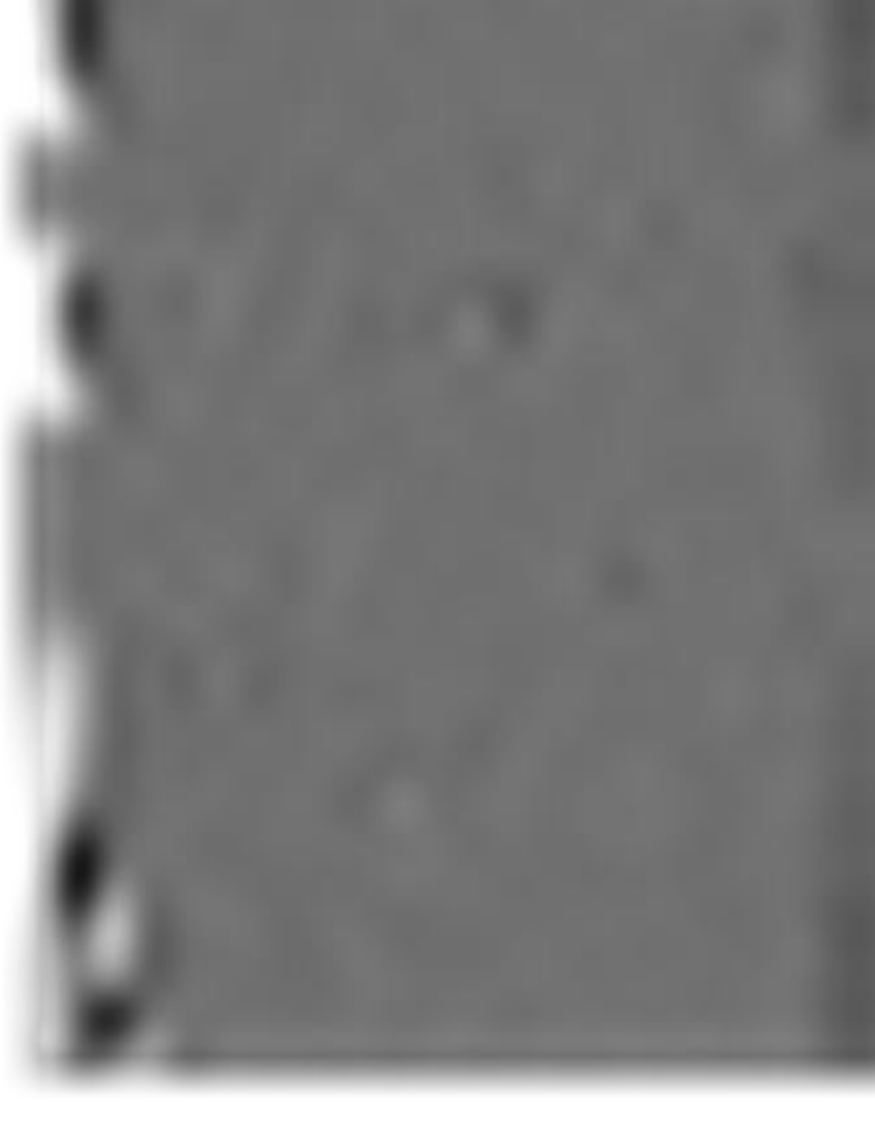}}
~\subfigure[]
{\label{fig:f}\includegraphics[width=8mm]{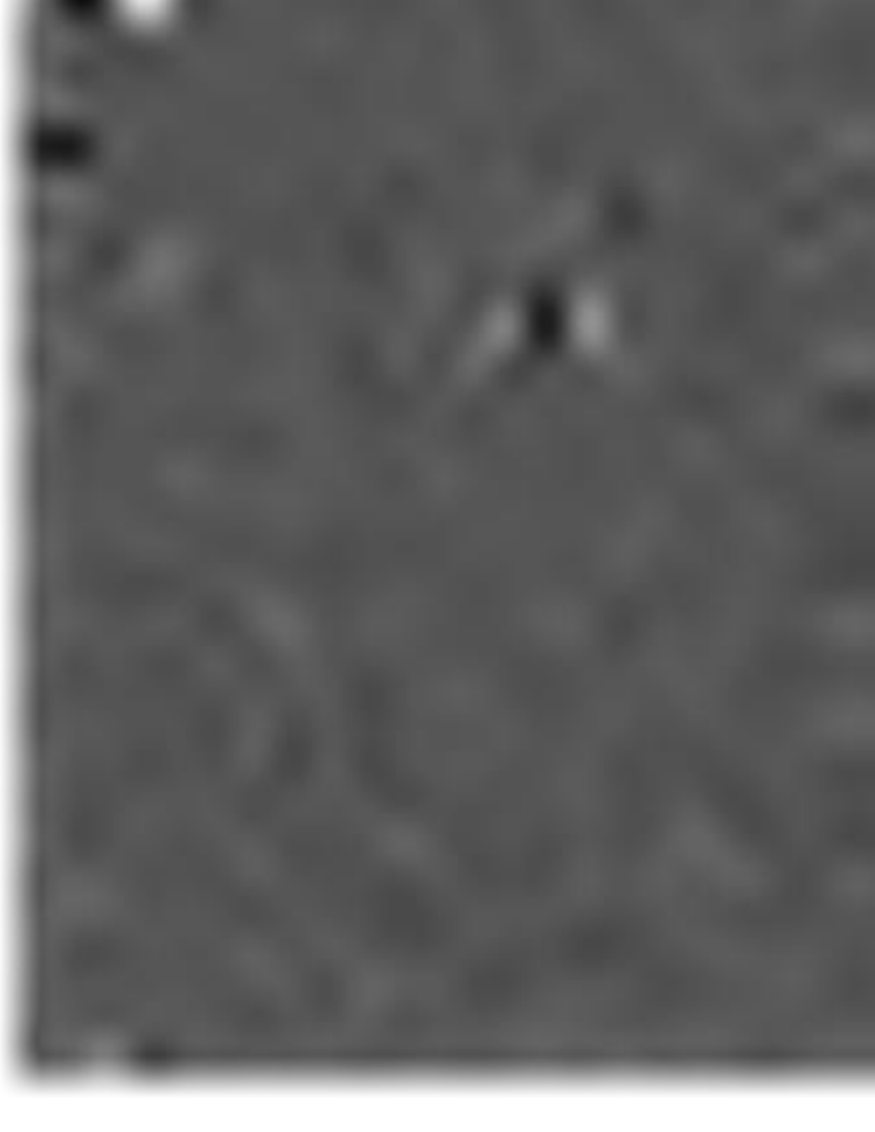}}
~\subfigure[]
{\label{fig:g}\includegraphics[width=25mm]{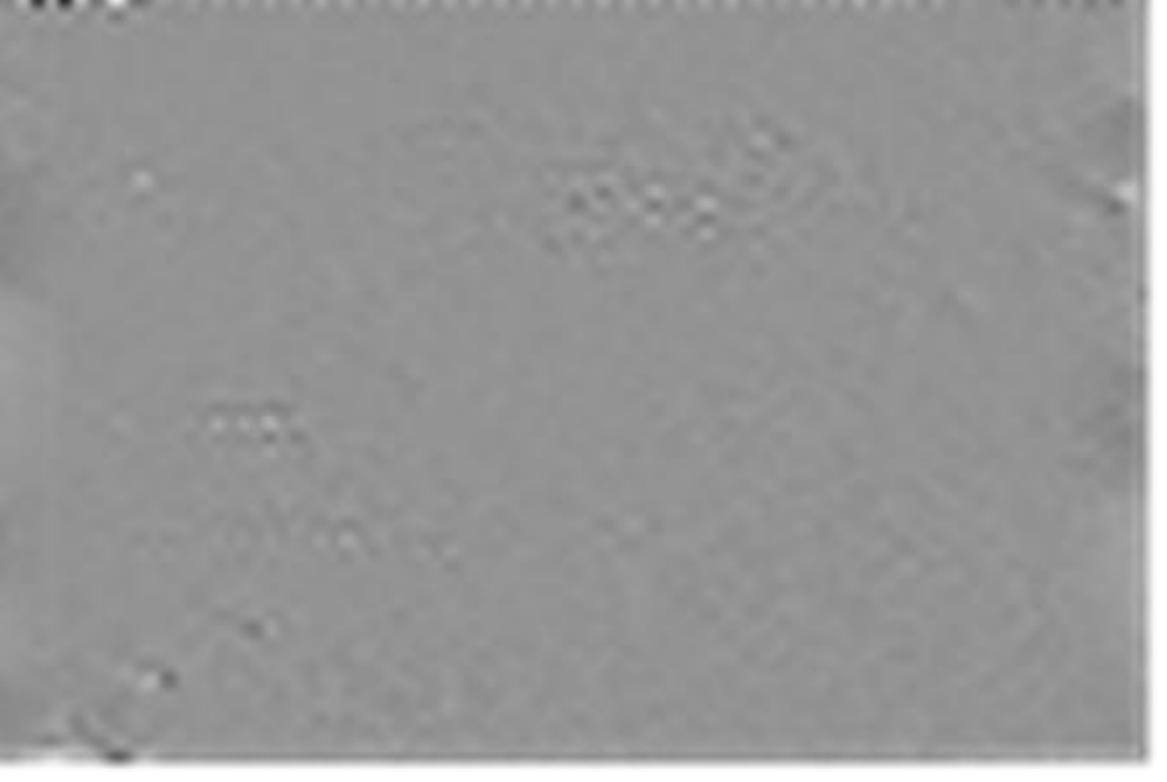}}
\caption{(a) a typical gray-scale image, (b)-(g) sub-bands obtained after contourlet transform.}
\label{fig:sub}
\end{figure}

We utilized MATLAB\textsuperscript{\textregistered} software to evaluate the performance of our method. At first, we cropped the face area from input image manually. Then, all images are resized to $120\times90$ pixels. The $fa$ and $fb$ indices are considered as one class. So, we have 7 pose classes. From each pose of database we randomly select 150 images (totally 1050 images). From each pose, we choose 10 images as train images and the other remained images are considered as test images.

\subsection{Feature Classification }
\begin{figure}[h!]
\centering
\includegraphics[width=90mm]{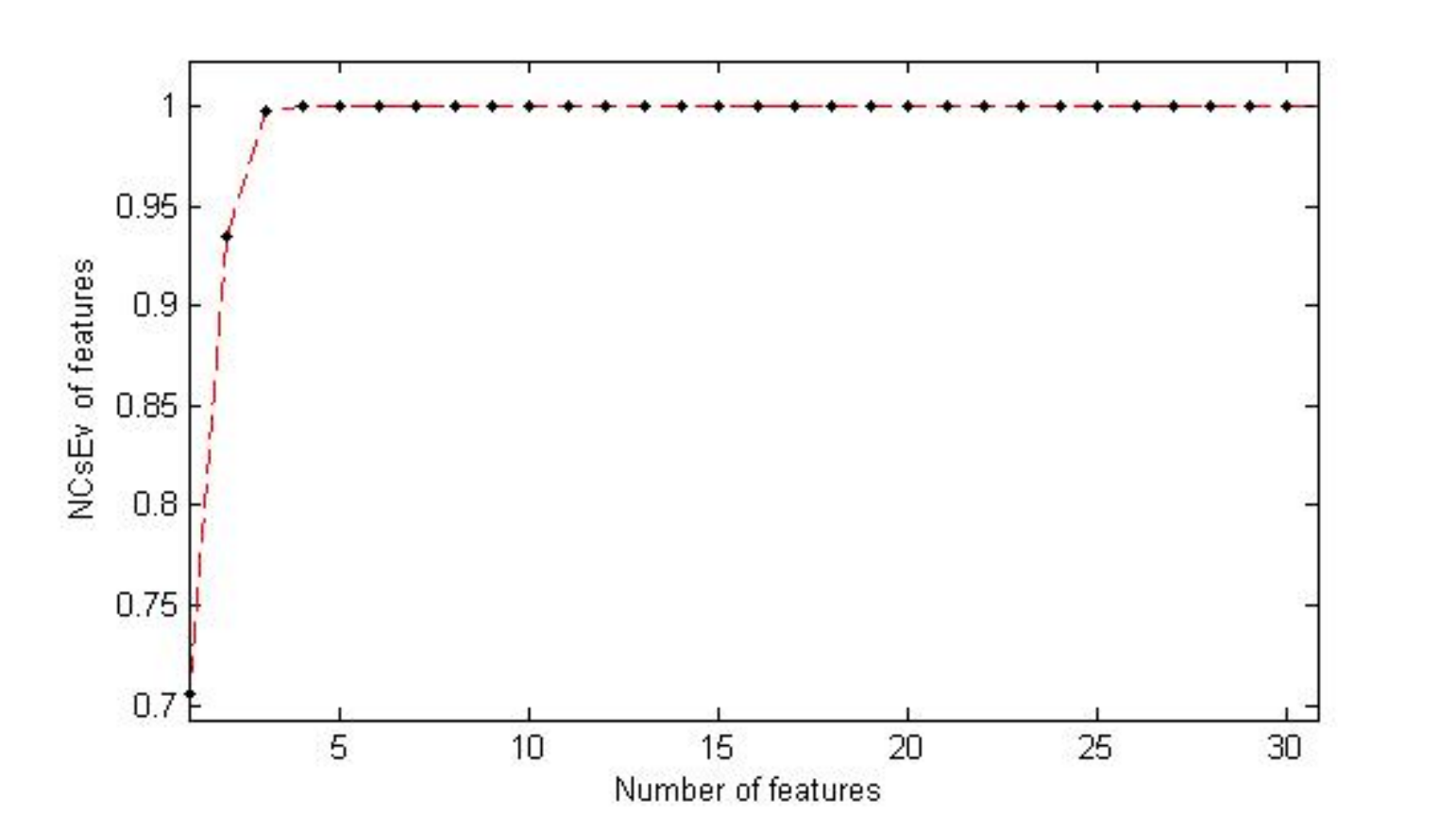}
\caption{Normalized cumulative summation of eigenvalues versus number of features.}
\label{fig:NCsEv}
\end{figure}

\begin{figure}[h]
\centering
\includegraphics[width=90mm]{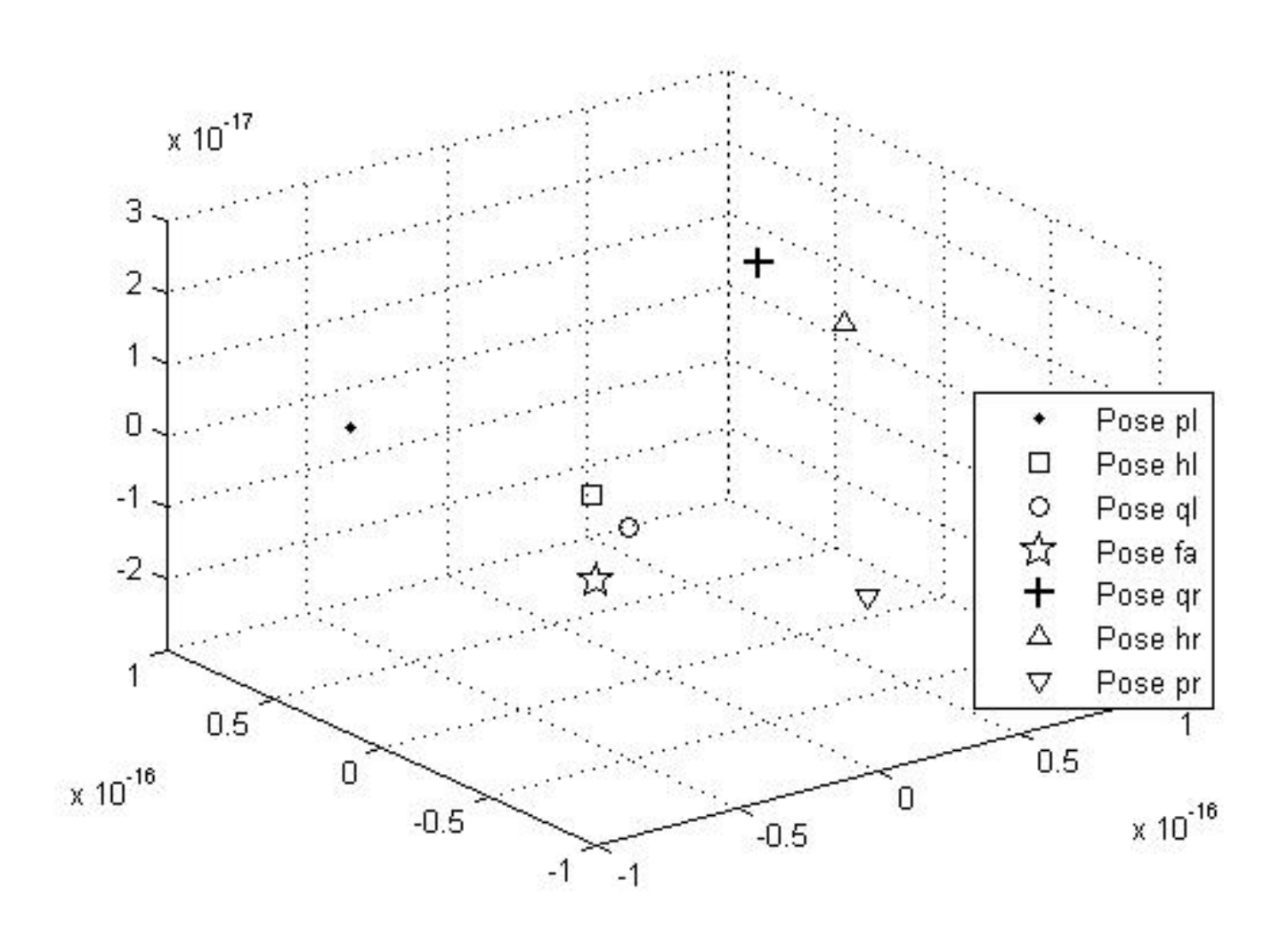}
\caption{Scatter plot of features obtained by PCA and LDA.}
\label{fig:scatter}
\end{figure}

 Among the 6 sub-bands obtained from contourlet transform as shown in Fig. \ref{fig:sub}, we use those sub-bands that labeled as $b$, $c$, $e$, and $f$ and are $30\times23$ pixels. Also, sub-bands $d$ and $g$ are $60\times90$ pixels. We consider the value of each pixel as one feature. Therefore, the dimension of feature space is $30\times23\times4$ which is equal to 2760. This dimensionality is high and classifying such large number of features requires high complexity. Also, all of these features are not proper and many of them are redundant. As mentioned previously, in order to remove the redundancy from feature space, PCA and LDA are used. For this purpose, the normalized cumulative summation of eigenvalue corresponding to $i$th feature is calculated as follows:

\begin{equation}
NCsEv(i) = \frac{\sum_{n=1}^{i}{Ev(n)}}{\sum_{n=1}^{M}{Ev(n)}} \quad\quad i = 1, ..., M
\label{1}
\end{equation}
where $Ev(x)$ is the eigenvalue of the $n$th feature, and $M$ is the total number of features. $NCsEv$ is shown Fig. \ref{fig:NCsEv}. According to $NCsEv(i)$ of features, we keep 3 features and remove others. In order to have a better representation, we only show $NCsEv$ for features 1 to 31. For features from 32 to 2760, $NCsEv$ is equal to 1. The scatter plots of the features obtained by PCA and LDA are shown in Fig. \ref{fig:scatter}. Scatter plot shows the values of features for different classes. It is clear that the proposed method has good ability in separating different pose classes.

In Table \ref{tab:confusion1} and Table \ref{tab:confusion2}, the confusion matrices of proposed method after applying k-NN and minimum distance classifiers are presented, respectively. It is clear that proposed method can classify different poses truly (i.e.: 100\% classification accuracy).

\begin{table}[h]
\caption{Confusion matrix for k-NN classifier.}
\centering
\scalebox{1.2}{
\begin{tabular}{cc|c|c|c|c|c|c|c|l}
\cline{3-9}
& & \multicolumn{7}{|c|}{Output Class} \\
\cline{3-9}
& & $pl$ & $hl$ & $ql$ & $fa$ & $qr$ & $hr$ & $pr$ \\
\cline{1-9}
\multicolumn{1}{|c|}{\multirow{4}{2mm}{\begin{sideways}\centering\parbox{19mm}{Target Class}\end{sideways}}} &
\multicolumn{1}{|c|}{$pl$} & 140 & 0 & 0 & 0 & 0 & 0 & 0 \\ \cline{2-9}
\multicolumn{1}{|c|}{}                        &
\multicolumn{1}{|c|}{$hl$} & 0 & 140 & 0 & 0 & 0 & 0 & 0 \\ \cline{2-9}
\multicolumn{1}{|c|}{}                        &
\multicolumn{1}{|c|}{$ql$} & 0 & 0 & 140 & 0 & 0 & 0 & 0 \\ \cline{2-9}
\multicolumn{1}{|c|}{}                        &
\multicolumn{1}{|c|}{$fa$} & 0 & 0 & 0 & 140 & 0 & 0 & 0 \\ \cline{2-9}
\multicolumn{1}{|c|}{}                        &
\multicolumn{1}{|c|}{$qr$} & 0 & 0 & 0 & 0 & 140 & 0 & 0 \\ \cline{2-9}
\multicolumn{1}{|c|}{}                        &
\multicolumn{1}{|c|}{$hr$} & 0 & 0 & 0 & 0 & 0 & 140 & 0 \\ \cline{2-9}
\multicolumn{1}{|c|}{}                        &
\multicolumn{1}{|c|}{$pr$} & 0 & 0 & 0 & 0 & 0 & 0 & 140 \\ \cline{1-9}
\end{tabular}}
\label{tab:confusion1}
\end{table}

\begin{table}[h]
\caption{Confusion matrix for minimum distance classifier.}
\centering
\scalebox{1.2}{
\begin{tabular}{cc|c|c|c|c|c|c|c|l}
\cline{3-9}
& & \multicolumn{7}{|c|}{Output Class} \\
\cline{3-9}
& & $pl$ & $hl$ & $ql$ & $fa$ & $qr$ & $hr$ & $pr$ \\
\cline{1-9}
\multicolumn{1}{|c|}{\multirow{4}{2mm}{\begin{sideways}\centering\parbox{19mm}{Target Class}\end{sideways}}} &
\multicolumn{1}{|c|}{$pl$} & 140 & 0 & 0 & 0 & 0 & 0 & 0 \\ \cline{2-9}
\multicolumn{1}{|c|}{}                        &
\multicolumn{1}{|c|}{$hl$} & 0 & 140 & 0 & 0 & 0 & 0 & 0 \\ \cline{2-9}
\multicolumn{1}{|c|}{}                        &
\multicolumn{1}{|c|}{$ql$} & 0 & 0 & 140 & 0 & 0 & 0 & 0 \\ \cline{2-9}
\multicolumn{1}{|c|}{}                        &
\multicolumn{1}{|c|}{$fa$} & 0 & 0 & 0 & 140 & 0 & 0 & 0 \\ \cline{2-9}
\multicolumn{1}{|c|}{}                        &
\multicolumn{1}{|c|}{$qr$} & 0 & 0 & 0 & 0 & 140 & 0 & 0 \\ \cline{2-9}
\multicolumn{1}{|c|}{}                        &
\multicolumn{1}{|c|}{$hr$} & 0 & 0 & 0 & 0 & 0 & 140 & 0 \\ \cline{2-9}
\multicolumn{1}{|c|}{}                        &
\multicolumn{1}{|c|}{$pr$} & 0 & 0 & 0 & 0 & 0 & 0 & 140 \\ \cline{1-9}
\end{tabular}}
\label{tab:confusion2}
\end{table}

Islam $et\ al.$ \cite{Islam} used bilateral filtering and wavelet transform in pre-processing step. They used isometric projection based subspace learning for extraction of discriminant feature vectors. They chose 5000 images from FERET database, but used five different poses, as $pl$, $ql$, $fa$, $qr$, and $ql$ face representations. PCA, LDA, and isometric projection (IsoP) are used separately for dimensionality reduction. For classification of feature vectors, they used the k-NN classifier. They achieved 90.338\%, 95.22\%, and 97.07\% recognition rates for PCA, LDA, and IsoP approaches, respectively. Comparison between the results of our method and the best results of \cite{Islam} is presented in Table \ref{tab:comparison}. From Table \ref{tab:comparison}, it is obvious that the proposed method has better accuracy than \cite{Islam} even in the case that we use 7 classes.

\section{Conclusion}\label{Conlusion}
In this paper we proposed a novel method for head pose estimation with less amount of features. In feature extraction step, we used contourlet transform. Then, we used PCA and LDA approaches for reduction of the dimensionality of the feature space. To evaluate the proposed algorithm, 1050 images from seven different poses of FERET database were selected. k-NN and minimum distance classifiers were used to classify the different poses. Experimental results demonstrate that the proposed algorithm achieve 100\% classification rate on images selected from FERET database.

\begin{table}[h!]
\caption{Performance comparison of different methods.}
\centering
\scalebox{1.2}{
\begin{tabular}{|c|p{1.5cm}|p{1.5cm}|p{1.5cm}|}
  \hline
  \textbf{Method} & Proposed method with 7 classes & Proposed method with 4 classes & Method of \cite{Islam} with 5 classes \\ \hline
  \textbf{Accuracy} & 100\% & 100\% & 97.07\% \\
  \hline
\end{tabular}}
\label{tab:comparison}
\end{table}

\nocite{*}
\bibliographystyle{IEEEtran}
\bibliography{pose}

\end{document}